\documentclass[journal]{IEEEtran}

\usepackage[caption=false,font=footnotesize]{subfig}
\usepackage{cite}
\usepackage{amsmath,amssymb,amsfonts}
\usepackage{algorithmic}
\usepackage{graphicx}
\usepackage{textcomp}
\usepackage{xcolor}
\usepackage{url}
\usepackage{booktabs} 
\usepackage{multirow} 
\usepackage{tabularx}

\begin{document}

\title{Building UI/UX Dataset for Dark Pattern Detection and YOLOv12x-based Real-Time Object Recognition Detection System}

\author{
Se-Young~Jang\textsuperscript{1},
Su-Yeon~Yoon\textsuperscript{2},
Jae-Woong~Jung\textsuperscript{3},
Dong-Hun~Lee\textsuperscript{3},
Seong-Hun~Choi\textsuperscript{4},
Soo-Kyung~Jun\textsuperscript{2},
Yu-Bin~Kim\textsuperscript{2},
Young-Seon~Ju\textsuperscript{5},
Kyounggon Kim\textsuperscript{6}\\
\vspace{1em} 
\textit{\textsuperscript{1}Baekseok University, Republic of Korea}\\
\textit{\textsuperscript{2}Ewha Womans University, Republic of Korea}\\
\textit{\textsuperscript{3}Tech University of Korea, Republic of Korea}\\
\textit{\textsuperscript{4}Best of the Best 14th Program, KITRI, Republic of Korea} \\
\textit{\textsuperscript{5}Korea Internet \& Security Agency (KISA), Republic of Korea}\\
\textit{\textsuperscript{6}Naif Arab University for Security Sciences, Saudi Arabia}
}

\markboth{}
{Jang \MakeLowercase{\textit{et al.}}: Building UI/UX Dataset for Dark Pattern Detection}

\maketitle

\begin{abstract}
With the accelerating pace of digital transformation and the widespread adoption of online platforms, both social and technical concerns regarding \emph{dark patterns}—user interface designs that undermine users’ ability to make informed and rational choices—have become increasingly prominent. As corporate online platforms grow more sophisticated in their design strategies, there is a pressing need for proactive and real-time detection technologies that go beyond the predominantly reactive approaches employed by regulatory authorities.

In this paper, we propose a visual dark pattern detection framework that improves both detection accuracy and real-time performance. To this end, we constructed a proprietary visual object detection dataset by manually collecting 4,066 UI/UX screenshots containing dark patterns from 194 websites across six major industrial sectors in South Korea and abroad. The collected images were annotated with five representative UI components commonly associated with dark patterns: \emph{Button}, \emph{Checkbox}, \emph{Input Field}, \emph{Pop-up}, and \emph{QR Code}. This dataset has been publicly released to support further research and development in the field.

To enable real-time detection, this study adopted the YOLOv12x object detection model and applied transfer learning to optimize its performance for visual dark pattern recognition. Experimental results demonstrate that the proposed approach achieves a high detection accuracy of 92.8\% in terms of mAP@50, while maintaining a real-time inference speed of 40.5 frames per second (FPS), confirming its effectiveness for practical deployment in online environments. Furthermore, to facilitate future research and contribute to technological advancements, the dataset constructed in this study has been made publicly available at \\https://github.com/B4E2/B4E2-DarkPattern-YOLO-DataSet.
\end{abstract}


\IEEEpeerreviewmaketitle

\section{Introduction}
\IEEEPARstart{W}{ith} the recent acceleration of digital transformation and the increasing complexity of online platforms, ``Dark Patterns'' have emerged as a major social and technical issue [2,3,4,5]. These are deceptive practices that infringe upon users' rights to make rational choices by inducing irrational spending and errors—for instance, ``Misdirection,'' where a subscription cancellation button is displayed as small and gray, while the button to maintain the subscription is emphasized with large, conspicuous colors. Although legal, policy, and technical efforts are underway globally to regulate these practices, the deceptive design strategies employed by corporations are simultaneously becoming more sophisticated. Consequently, there is an imperative need for proactive detection technologies that operate in real-time, transcending the reactive measures typically taken by regulatory authorities.

The severity of dark patterns has been corroborated by official legislative amendments and the establishment of guidelines within the international community, thereby substantiating the necessity for technical countermeasures [3,4,5]. For instance, the European Union (EU) has prohibited the use of dark patterns through the Digital Services Act (DSA) [3], and in South Korea, relevant authorities such as the Korea Fair Trade Commission (KFTC) are actively implementing response measures, including defining categories of dark patterns and publishing regulatory guidelines [4,14]. However, despite dark patterns emerging as a critical social issue, there remains a significant deficiency in visual datasets and detection technologies capable of automatically identifying these patterns to enhance regulatory effectiveness [2,5,7,12].

This paper proposes a methodology designed to enhance both the accuracy and real-time capability of visual dark pattern detection. To this end, our approach centers on a manually constructed dataset and the YOLO object detection model, implemented as follows. Regarding the dataset construction, we manually collected 4,066 UI/UX screenshots exhibiting dark patterns by analyzing 194 websites across six major industrial sectors—E-commerce, Finance, Travel \& Hospitality, Media \& Content, Sharing Economy \& Administration, and Search Portals/Press—both within South Korea and internationally. This extensive collection was aimed at encompassing the diverse typologies of dark patterns. Consequently, this dataset establishes a new benchmark for research in visual dark pattern detection and serves as a foundational resource for model training.

Next, utilizing the collected dataset, we present a new standard for technology capable of rapidly and accurately classifying and detecting dark pattern UI objects within images in real-time. This is achieved through the application of the YOLO (You Only Look Once) object detection model, which is optimally suited for real-time detection goals.
The remainder of this paper is organized as follows. Section II reviews the evolution of the YOLO series and provides a detailed investigation of dark pattern types. Section III describes the construction of the proprietary UI/UX dataset, including the collection and labeling processes. Section IV explains the rationale for selecting the YOLOv12x model and the methodology for model enhancement through transfer learning. Section V presents the experimental results and a comprehensive performance evaluation. Finally, Section VI concludes the paper and discusses future research directions.

\vspace{0.3cm}

\section{Literature Review}

\subsection*{Investigation of Dark Pattern Types}
With the recent rapid expansion of the online platform and mobile application markets, the design strategies of dark patterns intended to induce irrational user choices have become increasingly sophisticated. Notably, according to a large-scale empirical study on mobile applications, 95\% of the examined apps contained at least one dark pattern [2,4,5], with an average of approximately 7.4 deceptive designs identified per app. Of particular significance is the finding that ``Interface Interference''—a type that manipulates visual hierarchy to induce specific choices—was present in 61\% of the total apps. Furthermore, the ``Nagging'' type, which continuously interrupts user tasks via pop-ups and banners, also accounted for a substantial proportion at 55\% [2,4,5]. This demonstrates that dark patterns have evolved beyond issues of mere textual manipulation to deeply infiltrate the entire spectrum of visual UI/UX design, including screen composition and layout [15,16].

Furthermore, analyses of consumer harm cases, both domestic and international, reveal this trend. According to a survey on dark pattern harm experiences among the general public, the most frequently experienced type was ``Hard to Cancel'' (34\%), which complicates the subscription cancellation process to prevent user churn. This was followed by ``Hidden Costs'' (21\%), which conceals or visually excludes price and option information within the screen [4,5]. These types are closely linked to techniques that distort the user's decision-making process through the placement of UI/UX elements, visual information, and information concealment, such as ``Interface Interference'' or ``Obstruction.'' In other words, this can be viewed as an instance where the domain of interaction design—originally intended to make behavioral flows and device responses natural and convenient during service usage—is being exploited [2,4,5].

However, existing technical detection studies [17,18] have primarily focused on methods that extract text from web pages and analyze it using Natural Language Processing (NLP) models [7,12]. This approach has been identified as having a fundamental limitation: it fails to identify dark patterns composed of images or graphic elements [7,12]. Considering that many recent websites implement dark patterns by modifying visual attributes—such as image banners and the color, size, and placement of buttons—rather than relying on text, relying solely on text-centric analysis may lead to False Negatives (FN), where actual dark patterns exist but remain undetected and are judged as benign [19].

Therefore, comprehensively considering that the majority of dark patterns undermining user experience stem from visual UI/UX elements, and acknowledging the limitations of existing text-based detection technologies, it is imperative to analyze the UI/UX screens themselves to effectively detect dark patterns in real-time. Accordingly, this study determined that it is essential to collect actual UI/UX screens where dark patterns are manifested and to construct a dataset optimized for visual object detection.

\section{Dark Pattern Dataset}

In this study, we constructed a UI/UX dataset to maximize the training performance of visual object detection models and to implement a generalized dark pattern detection technology that is unbiased towards specific domains[Table I]. Given that the visual manifestations of dark patterns (e.g., layout, color, and typography) vary depending on the nature and purpose of the service, the scope of data collection was designed to broadly encompass major online platforms both in South Korea and overseas, rather than being confined to a single industrial sector.

\begin{table}[htbp]
\caption{Target Platforms for Data Collection by Industrial Sector}
\label{tab:target_platforms}
\centering
\begin{tabular}{|c|l|p{4cm}|}
\hline
\textbf{No.} & \textbf{Industry(count)} & \textbf{Target Platforms} \\ \hline
1 & E-commerce(1,706) & Comprehensive Shopping Malls, Fashion, Food \& Groceries, Cross-border Shopping Platforms \\ \hline
2 & Travel \& Acc(750) & Yanolja, Yeogi Eottae, Agoda, Airbnb, Skyscanner \\ \hline
3 & Finance(400) & Commercial Banks, Fintech \\ \hline
4 & Media \& Content(610) & OTT Services (Netflix, etc.), Music Streaming \\ \hline
5 & Public(350) & National Tax Service Hometax, Gov 24 \\ \hline
6 & Press(250) & The Chosun Daily, JoongAng Daily \\ \hline
\end{tabular}
\end{table}

\vspace{0.3cm}

The targets for data collection were primarily selected from websites and mobile applications with high usage frequency in South Korea. The specific collection targets and their characteristics are detailed as listed in  Table I.

Items 1 through 6 provide a detailed explanation of following.
First, the E-commerce sector, identified as the domain where dark patterns manifest most frequently, was allocated the largest proportion of the dataset. This selection encompasses a wide range of platforms, including comprehensive shopping malls (e.g., Coupang, Gmarket, 11Street), fashion platforms (e.g., Musinsa, Zigzag, Ably), grocery services (e.g., Market Kurly), home shopping channels (e.g., GS SHOP, Lotte Homeshopping), and cross-border shopping platforms (e.g., Amazon, Shein, Qoo10). The primary objective of this concentration was to intensively secure visual patterns designed to induce purchases, such as ``Low Stock'' (imminent sell-out), ``Limited Time'' constraints, and ``Hidden Costs.'' Consequently, we collected a total of 1,706 images in this sector.

Second, for the Travel \& Accommodation sector, we selected booking intermediary platforms such as Yanolja, Yeogi Eottae, Agoda, Airbnb, and Skyscanner. These platforms provide critical data for analyzing how ``Scarcity'' and ``Urgency'' patterns such as ``real-time booking status'' and ``imminent deadlines'' are visually implemented within the interface. For this sector, we gathered 750 images.

Third, for the Finance sector, we included major commercial banks (e.g., KB Kookmin, Shinhan, Woori), fintech platforms (e.g., Toss, KakaoBank), and credit card companies. We compiled 400 images from these financial services.

Fourth, the Media \& Content sector was included, comprising OTT services (e.g., Netflix, Tving, Wavve), music streaming platforms (e.g., Melon, Bugs), and webtoon services. In this domain, the primary focus was placed on collecting UI/UX patterns related to ``Hard to Cancel'' (subscription cancellation obstruction) and ``Forced Action'' strategies. A total of 610 images were collected from this domain.

Fifth, the Public and Administration sector was included to secure a control group and benign UI data. For this purpose, we collected data from platforms such as the National Tax Service Hometax, Government24, and various local government websites. We secured 350 images from this sector.

Sixth, the Press sector included major news outlets (e.g., Chosun Ilbo, JoongAng Ilbo). This inclusion was designed to enable the model to learn the visual distinctions between interfaces lacking explicit commercial intent and commercial interfaces containing dark patterns. We acquired 250 images from news outlets.

\begin{figure}[htbp]
    \centering

    \includegraphics[width=0.9\linewidth]{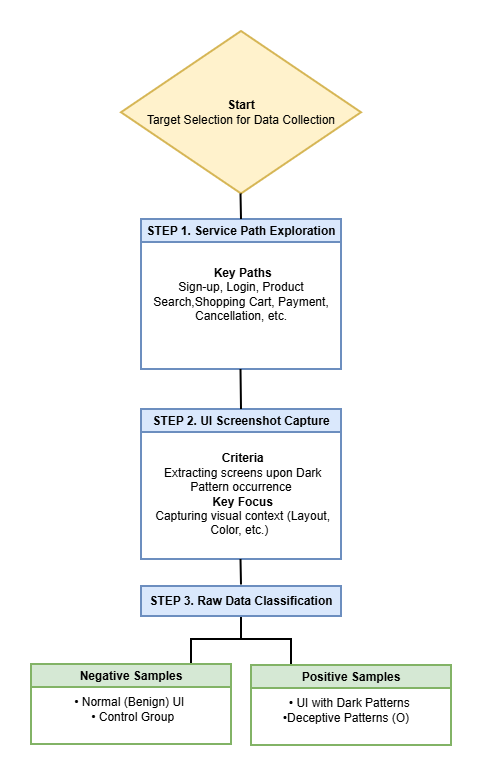}

    \caption{Collection of Raw Data on Dark Patterns}

    \label{fig:collection_process}
\end{figure}

In this study, raw data was obtained by capturing UI/UX elements suspected of dark patterns from the web and mobile interfaces of the 194 selected platforms. The data collection process followed the procedures outlined in Figure 1.

\vspace{0.3cm}

\subsubsection{Step 1. Service Path Exploration}
Key paths with a high probability of exposing users to dark patterns were directly navigated on each platform. These processes included account registration, log-in, product search, adding to cart, payment, and subscription cancellation.

\subsubsection{Step 2. UI Screenshot Capture}
We captured screens from web and mobile platforms where dark patterns clearly appeared. The focus was to obtain UI/UX images that contain the full visual context—such as button size/location, popup layout, and color contrast—rather than extracting text alone.

\subsubsection{Step 3. Raw Data Classification}
The raw dataset was constructed by classifying the captured screenshots into Positive Samples (containing dark patterns) and Negative Samples (control group, containing standard UI/UX elements).

Through this process, a total of 4,066 UI/UX images were secured, successfully laying the groundwork for training object detection models based on visual features, rather than text-centric elements in Figure 2.

\subsection{Dataset Type Classification and Labeling}
The raw dataset, comprising a total of 4,066 images collected from the 194 platforms across the six major industries described earlier, consists of actual service screens containing dark patterns. In this study, we performed a classification and labeling process on the UI/UX elements to enable the object detection model to accurately identify visual objects associated with dark patterns within these images.


Dark patterns are not manifested by specific phrases or images in isolation; rather, they are implemented through the combination and arrangement of fundamental UI/UX components—such as buttons, pop-up windows, and input fields—and the modification of visual attributes (e.g., size and color). Therefore, for the model to effectively detect dark patterns, it is essential to first accurately recognize the basic UI/UX elements constituting the screen.


To this end, this study utilized the Roboflow platform to classify all interactive UI/UX elements within the collected images into five key classes: \emph{button}, \emph{checkbox}, \emph{popup}, \emph{input\_field}, and \emph{qr\_code}. Based on this classification, bounding boxes were manually annotated for the corresponding UI/UX elements.

\begin{figure*}[t]
  \centering
  
  \setlength{\tabcolsep}{10pt} 
  \renewcommand{\arraystretch}{1.5}
  
  \begin{tabular}{|c|c|}
    \hline

    \includegraphics[width=0.45\linewidth, keepaspectratio]{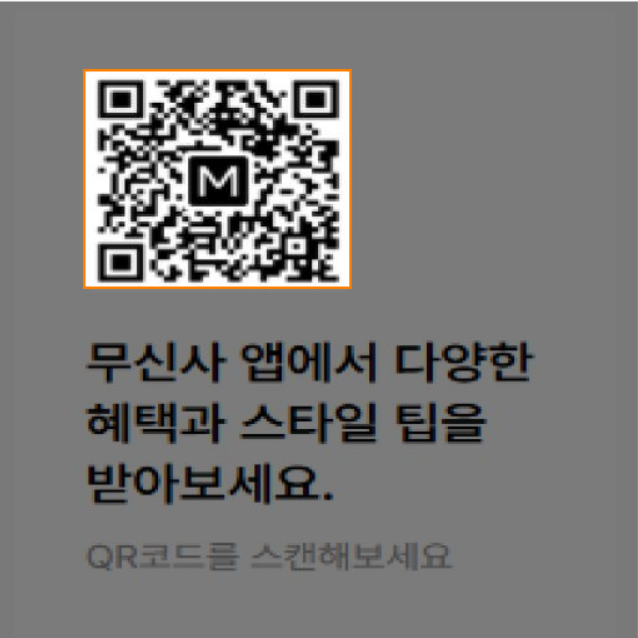} & 
    \includegraphics[width=0.45\linewidth, keepaspectratio]{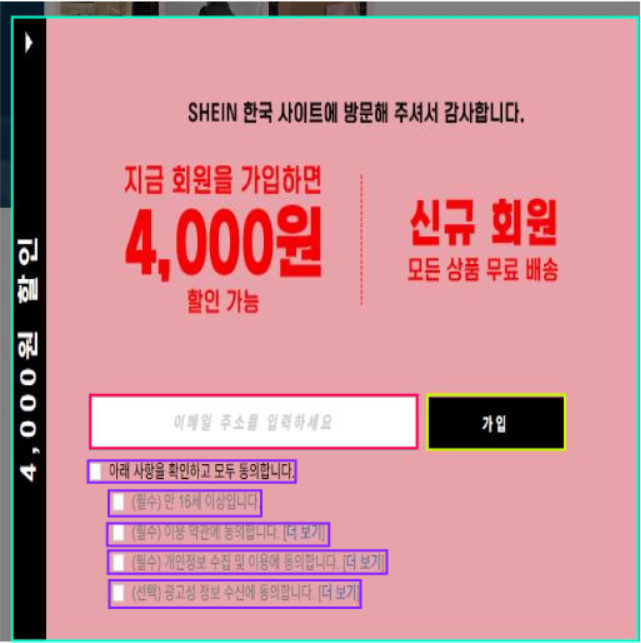} \\
    \hline

    (a) Collection Process & (b) Labeled Examples \\
    \hline
  \end{tabular}
  
  \caption{Overview of Data Collection Process (Left) and Labeled Dataset Examples (Right)}
  \label{fig:combined_images_table_large}
\end{figure*}

To minimize errors and ensure data reliability, cross-validation was performed by at least two labelers on the entire dataset. This labeled dataset plays a pivotal role in enabling the YOLO model to learn the location and shape of objects constituting dark patterns across various web environments. The details of the five key classes are summarized in Table II.

\begin{table}[htbp]
\caption{Five Key Detection Classes and Dataset Counts}
\label{tab:classes}
\centering
\begin{tabular}{|l|p{5cm}|c|}
\hline
\textbf{Class} & \textbf{Overview and Relevance to Dark Patterns} & \textbf{Total} \\ \hline
Button & A class encompassing all button elements designed to induce user clicks, such as ``Purchase'', ``Cancel'', and ``Close''. & 1,691 \\ \hline
Checkbox & A class associated with selection options such as ``Agree'' and ``Opt-out''. & 1,045 \\ \hline
Input Field & A class representing elements where users input text, such as search bars and login fields. & 1,537 \\ \hline
Popup & A class containing advertisements and notification windows that appear unexpectedly at the center or bottom of the screen. & 210 \\ \hline
QR Code & A class utilized for redirection, such as prompting mobile app installation or automatically navigating users to a new specific page. & 430 \\ \hline
\end{tabular}
\end{table}


\subsection{Rationale for Selecting the Object Detection Model}
With advancements in deep learning, the field of object detection has evolved to automatically learn hierarchical features from data based on Convolutional Neural Networks (CNN) [10, 11]. Modern object detection models are generally categorized into Two-Stage Detectors, which sequentially perform region proposal and classification, and One-Stage Detectors, which integrate these processes into a single step [10, 11].

Consequently, considering the environmental characteristics where dark patterns occur, this study prioritized real-time processing speed alongside detection accuracy as the key criteria for model selection.

The R-CNN family, representing early two-stage detectors, demonstrated high accuracy but is known to face limitations regarding computational speed [12, 14]. Although Faster R-CNN was proposed to address this by introducing a Region Proposal Network (RPN) to enhance speed through GPU computation [13], it retains a two-stage structure where region proposal and object classification are separated. Consequently, bottlenecks in computational cost persist in environments requiring real-time performance [12, 14]. In particular, minimizing latency is essential for detecting dark patterns in environments where user interactions occur instantaneously, such as web browsing.

In contrast, the YOLO (You Only Look Once) algorithm overcame these structural limitations by redefining object detection as a single regression problem [12]. YOLO divides the input image into a grid and utilizes a single neural network to simultaneously predict bounding box coordinates and class probabilities [12]. This one-stage detection approach is effective in reducing false positives by capturing the global context of the image at once. Most importantly, it demonstrates superior performance in terms of inference speed by eliminating the need for complex pipelines [12, 14].

Notably, Nguyen Yinkfu et al [20], comparative experiments between YOLO and Faster R-CNN in environments requiring immediate decision-making, such as autonomous driving, revealed distinct characteristics. While Faster R-CNN demonstrates high precision for small objects, it incurs a heavy computational load. In contrast, YOLO (particularly models post-v5) was found to offer the most optimal speed-accuracy trade-off [14], achieving comparable accuracy while maintaining speeds suitable for real-time processing.

Therefore, considering the system's objective to detect visually hidden or manipulated UI/UX elements in real-time without latency while users navigate, this study adopted the YOLO model as the core detection algorithm.

\subsection{YOLO (You Only Look Once)}

\subsubsection*{Basic Principles and Architecture of YOLO}

adopted in this study for real-time dark pattern detection, is a one-stage detector that defines object detection as a single regression problem, rather than a complex processing pipeline. By `looking once' at the input image to simultaneously predict object locations and classes, it ensures superior inference speed compared to traditional R-CNN-based models, making it optimized for real-time detection [6, 8, 11].

The core algorithm of YOLO begins by dividing the input image into an $S \times S$ grid. If the center of an object falls into a specific grid cell, that cell is responsible for detecting the object. Each grid cell predicts $B$ bounding boxes, the confidence scores for those boxes, and $C$ class probabilities [8, 11]. In this context, the confidence score is calculated as the probability of an object's presence, $Pr(Object)$, multiplied by the Intersection over Union (IoU)(1), which indicates how well the predicted box matches the ground truth (1).

\begin{equation}
C_{s} = Pr(Object) \times IoU_{pred}^{truth}
\label{eq:confidence}
\end{equation}

Training the YOLO network involves minimizing a multi-part loss function that aggregates localization loss, confidence loss, and classification loss. The loss function proposed in the original YOLO paper is defined as follows(2).

\begin{equation}
\begin{aligned}
Loss &= \lambda_{coord}\sum_{i=0}^{S^{2}}\sum_{j=0}^{B}\mathbb{1}_{ij}^{obj}[(x_{i}-\hat{x}_{i})^{2}+(y_{i}-\hat{y}_{i})^{2}] \\
&+ \lambda_{coord}\sum_{i=0}^{S^{2}}\sum_{j=0}^{B}\mathbb{1}_{ij}^{obj}[(\sqrt{w_{i}}-\sqrt{\hat{w}_{i}})^{2}+(\sqrt{h_{i}}-\sqrt{\hat{h}_{i}})^{2}] \\
&+ \sum_{i=0}^{S^{2}}\sum_{j=0}^{B}\mathbb{1}_{ij}^{obj}(C_{i}-\hat{C}_{i})^{2} \\
&+ \lambda_{noobj}\sum_{i=0}^{S^{2}}\sum_{j=0}^{B}\mathbb{1}_{ij}^{noobj}(C_{i}-\hat{C}_{i})^{2} \\
&+ \sum_{i=0}^{S^{2}}\mathbb{1}_{i}^{obj}\sum_{c \in classes}(p_{i}(c)-\hat{p}_{i}(c))^{2}
\end{aligned}
\label{eq:loss}
\end{equation}

In this context, $\mathbb{1}_{ij}^{obj}$ signifies that the $j$-th bounding box predictor in cell $i$ is responsible for predicting the object. By assigning a higher weight $\lambda_{coord}$ to coordinate and size predictions when an object exists, the function reinforces localization accuracy. Conversely, it addresses training imbalance by reducing the weight for non-object backgrounds(2).


The inference process of YOLO can be summarized in the following pseudo-code in Figure 3.

\begin{figure}[htbp]
\centering
\begin{tabular}{|l|}
\hline
\textbf{Algorithm: YOLO Detection Pseudo-code} \\ \hline
\begin{minipage}{0.95\linewidth}
\vspace{0.1cm}

\scriptsize 
\begin{verbatim}
def yolo_detection(image, model, thresh):
    
    input_tensor = resize_and_normalize(
        image)
    predictions = model(input_tensor)
    
    boxes = []
    
    for grid_cell in predictions:
        if confidence > thresh:
            
            box = decode_coordinates(
                grid_cell)
            class_id = argmax(
                class_probabilities)
            
            boxes.append((box, class_id, 
                          confidence))
            
    final_detections = apply_nms(boxes)
    
    return final_detections
\end{verbatim}
\vspace{0.1cm}
\end{minipage} \\ \hline
\end{tabular}
\caption{YOLO Detection Pseudo-code}
\label{fig:yolo_pseudocode}
\end{figure}

\subsubsection*{Evolution of the YOLO Series and Model Scaling}
Since the release of its first version in 2016, YOLO has continuously improved its speed and accuracy, establishing itself as the de facto standard in the field of object detection [6, 10, 11].

While early models (V1-V3) secured real-time capabilities based on the Darknet architecture, they exhibited weaknesses in detecting small objects [6, 10]. To address this, YOLOv4 introduced the  Cross Stage Partial Network (CSPNet) backbone and the Mosaic data augmentation technique. Subsequently, YOLOv5, implemented in PyTorch, significantly improved user accessibility and ease of deployment, thereby establishing itself as an industry standard [6].

Subsequent models focused on optimizing structural efficiency and refining training strategies. YOLOv6 introduced a hardware-efficient design tailored for industrial applications [1, 6], while YOLOv7 maximized parameter efficiency via the E-ELAN architecture and trainable Bag-of-Freebies strategies [6]. YOLOv8 enhanced detection flexibility by adopting an anchor-free approach and a novel loss function, whereas YOLOv9 addressed the information loss issue in deep neural networks through PGI and the GELAN architecture [6]. Furthermore, YOLOv10 drastically reduced inference latency by eliminating the Non-Maximum Suppression (NMS) process [13].

Most recently, YOLOv11 improved feature extraction capabilities in various vision tasks by incorporating the C3k2 blocks and the C2PSA module [6]. YOLOv12, the latest iteration, introduces an `Attention-centric' design, surpassing the limitations of traditional CNN-based architectures. By integrating Area Attention with the R-ELAN backbone, it achieves real-time processing speeds comparable to CNNs while retaining the contextual understanding capabilities inherent to Transformer models [6].

In particular, to ensure flexibility across various hardware sets-ups, the YOLO series employs model scaling by modulating network depth and width. These models are generally categorized into five variants from Nano (N) to X-Large (X)—depending on their parameter counts and computational volume (FLOPs). The characteristics of each scale in YOLOv12 are summarized in Table III.

\begin{table}[t]
\caption{Performance Overview and Advantages/Disadvantages of the Five YOLOv12 Scales}
\label{tab:yolo_scales}
\centering
\scriptsize
\renewcommand{\arraystretch}{1.25}

\resizebox{\columnwidth}{!}{%
\begin{tabular}{|c|p{1.6cm}|p{1.6cm}|p{1.9cm}|}
\hline
\textbf{Scale} & \textbf{Key Features} & \textbf{Pros/Cons} & \textbf{Metrics (v12)} \\ \hline

Nano (N) &
Ultra-lightweight &
Ultra-low latency \newline Lower accuracy &
Params: 2.3M \newline Lat.: 1.84ms \\ \hline

Small (S) &
Speed/Acc balance &
Balanced performance &
Params: $>$N \newline Comp.: 1.5 \\ \hline

Medium (M) &
High accuracy &
Stable, slower than S &
Params: $>$S \newline Comp.: 2.0 \\ \hline

Large (L) &
Complex detection &
Good patterns \newline Slower &
Params: $>$M \newline Comp.: 2.5 \\ \hline

X-Large (X) &
Max depth/width &
Highest acc. \newline High GPU &
Params: 11.79M \newline mAP50: 55.2 \\ \hline

\end{tabular}
}
\end{table}

\subsection{Comparative Analysis and Final Model Selection}

\subsubsection*{Background on Model Selection}
Models ranging from YOLOv8 to YOLOv12 were selected as baselines to verify object detection performance. While early iterations (v1-v7) achieved innovations in speed, YOLOv8 established itself as the industry's de facto standard due to the stabilization of the anchor-free approach and the maximization of PyTorch-based flexibility [6, 9, 10]. Therefore, this study aims to identify the optimal model best suited for Web UI/UX environments by comparing YOLOv8 with YOLOv9, v10, v11, and the attention-based YOLOv12 in Table IV.

\subsubsection*{Comparison of Feature Extraction Capabilities}
This experiment was conducted as a preliminary study to verify the feature extraction capability of each model architecture and select the optimal model. We constructed a test dataset by extracting representative samples with the most distinct visual features for each class (Button, Checkbox, Input Field, Pop-up, QR code) from the collected data. We then evaluated how effectively the pre-trained weights of each model (based on the X scale) could recognize the UI objects.

The experimental results are presented in Table \ref{tab:model_comparison}. YOLOv8x demonstrated fast speed but occasionally missed fine-grained objects. YOLOv10x recorded the highest speed (52.1 FPS) but lacked precise feature extraction. Conversely, YOLOv12x, which possesses the highest number of parameters, generated the most sophisticated feature maps, recording 0.775 on the mAP@50-95 metric.

\begin{table}[htbp]
\caption{Performance Comparison of YOLO Series (X-Scale Models) Based on Core UI Object Feature Extraction}
\label{tab:model_comparison}
\centering
\begin{tabular}{|l|c|c|c|c|c|}
\hline
\textbf{Model} & \textbf{Prec.} & \textbf{Recall} & \textbf{mAP50} & \textbf{mAP50-95} & \textbf{FPS} \\ \hline
YOLOv8x & 0.885 & 0.872 & 0.901 & 0.715 & 45.2 \\ \hline
YOLOv9x & 0.898 & 0.889 & 0.918 & 0.732 & 38.5 \\ \hline
YOLOv10x & 0.891 & 0.884 & 0.912 & 0.728 & 52.1 \\ \hline
YOLOv11x & 0.908 & 0.899 & 0.930 & 0.751 & 42.8 \\ \hline
\textbf{YOLOv12x} & \textbf{0.927} & \textbf{0.918} & \textbf{0.948} & \textbf{0.775} & \textbf{40.5} \\ \hline
\end{tabular}
\end{table}

Therefore, to maximize the system's primary goals of `maximum accuracy' and `detection reliability', we adopted \textbf{YOLOv12x} as the final detection model in this study.

\section{YOLOv12x Model Enhancement and Evaluation}

\subsection{Transfer Learning and Training Strategy}
Based on the YOLOv12x model, we performed model enhancement using the 4,066 UI images. We adopted the Transfer Learning technique, where the pre-trained weights from general objects (such as the COCO Dataset) were set as initial values, followed by fine-tuning for the five classes specific to the web UI environment.

\subsection{Model Training Implementation}
The enhanced training of the YOLOv12x model was implemented in a Python environment utilizing the Ultralytics framework in Figure 4. The process was conducted on the Google Colaboratory (Colab) platform to ensure efficiency in data management and GPU resource utilization.

\begin{figure}[htbp]
\centering
\begin{tabular}{|l|}
\hline
\textbf{Implementation: YOLOv12x Optimization} \\ \hline
\begin{minipage}{0.95\linewidth}
\vspace{0.1cm}

\scriptsize 
\begin{verbatim}
from ultralytics import YOLO

# Load model
model = YOLO('/path/to/yolov12x/best.pt')

# Train model
results = model.train(
  
  data='/content/drive/MyDrive/UI_v3/'
       'data.yaml',
  epochs=100,
  imgsz=640,
  batch=16,
  
  name='yolov12x_custom_'
       'finetuning',
  optimizer='auto',
  exist_ok=True,
  plots=True
)
\end{verbatim}
\vspace{0.1cm}
\end{minipage} \\ \hline
\end{tabular}
\caption{Training Implementation Code}
\label{fig:training_code}
\end{figure}

\subsection{Final Performance Evaluation Results}
After completing 100 epochs of enhanced training, the final performance was evaluated using the validation set. The final training results show that the YOLOv12x model achieved a detection performance of \textbf{92.8\%} based on the overall class average mAP@50 as listed in Table V. This signifies that the model accurately detects dark pattern UI elements more than nine out of ten times in real-time web environments.

\begin{table}[htbp]
\caption{Final Performance Evaluation of the Enhanced YOLOv12x Model (Validation Results)}
\label{tab:final_results}
\centering
\begin{tabular}{|l|c|c|c|c|}
\hline
\textbf{Class} & \textbf{Precision} & \textbf{Recall} & \textbf{mAP@50} & \textbf{mAP@50-95} \\ \hline
\textbf{All} & \textbf{0.933} & \textbf{0.881} & \textbf{0.928} & \textbf{0.797} \\ \hline
QR Code & 0.987 & 0.998 & 0.995 & 0.862 \\ \hline
Checkbox & 0.963 & 0.941 & 0.977 & 0.770 \\ \hline
Popup & 0.948 & 0.877 & 0.948 & 0.940 \\ \hline
Button & 0.865 & 0.784 & 0.880 & 0.666 \\ \hline
Input Field & 0.971 & 0.732 & 0.838 & 0.749 \\ \hline
\end{tabular}
\end{table}

\section{Conclusion}
The purpose of this paper was to establish a technical basis for real-time dark pattern detection and user protection on online platforms. We achieved this by moving away from traditional text-centric detection and proposing a new approach in which the analysis is focused on the visual elements of the UI/UX directly.


The core achievement of this study was the establishment of a proprietary dataset by analyzing 194 platforms across six major domestic and international industry sectors. We collected 4,066 real service screens where dark patterns frequently occur and precisely labeled them into five key UI classes (Button, Checkbox, Input Field, Pop-up, QR code). One of the main objectives of this study was to make this dataset publicly available to support the research community. Consequently, this dataset was officially released globally through three major open-source platforms---Kaggle, Roboflow, and GitHub---to contribute to the advancement of the academic and technological ecosystem.

Furthermore, by enhancing the latest object detection model, YOLOv12x, through Transfer Learning on the aforementioned dataset, we achieved a remarkably high detection accuracy of 92.8\% based on mAP@50. Concurrently, by securing an inference speed of 40.5 FPS, we demonstrated the technical feasibility of identifying UI/UX elements in real-time without compromising the user experience in a web browsing environment.

In this study, we anticipate that this dataset will be widely utilized by subsequent researchers, thereby establishing a foundation that accelerates the development of dark pattern detection technology.


\end{document}